\documentclass[journal]{IEEEtran}

\usepackage[dvipsnames]{xcolor}
\usepackage{multirow}
\usepackage{subcaption}

\ifCLASSINFOpdf
  \usepackage[pdftex]{graphicx}
\usepackage{amsfonts}
\usepackage{amsmath}
  \graphicspath{{./figures/}}
  \DeclareGraphicsExtensions{.pdf,.jpeg,.png}
\else
\fi
\usepackage{soul}

\usepackage{amsmath}
\DeclareMathOperator*{\argmax}{arg\,max}

\usepackage{newtxmath}

\usepackage{amsfonts}
\usepackage{algorithmic}

\usepackage{xcolor}
\usepackage[linesnumbered,ruled,vlined]{algorithm2e}
\SetKwInput{KwInput}{Input}                
\SetKwInput{KwOutput}{Output}              

\usepackage{hyperref}
\usepackage{url}

\definecolor{Lime}{RGB}{102,255,0}

\begin{document}

%


\title{TDIOT: Target-driven Inference for Deep Video Object Tracking}
%
%
%

\author{\IEEEauthorblockN{Filiz~Gurkan,  Llukman~Cerkezi, Ozgun~Cirakman and~Bilge~Gunsel
}

\thanks{F.Gurkan, L. Cerkezi, O.Cirakman and B.Gunsel are with Multimedia Signal Proc. and Pattern Recognition Group, Departmant of EE, Istanbul Technical University, Turkey (e-mail: \{gurkanf,cirakmano,gunselb\}@itu.edu.tr)}

}

%
%

\markboth{}%
{ \MakeLowercase{\textit{et al.}}:TDIOT}
%



\maketitle

\begin{abstract}

Recent tracking-by-detection approaches use deep object detectors as target detection baseline, because of their high performance on still images. For effective video object tracking, object detection is integrated with a data association step performed by either a custom design inference architecture or an end-to-end joint training for tracking purpose.
In this work, we adopt the former approach and use the pre-trained Mask R-CNN deep object detector as the baseline. We introduce a novel inference architecture placed on top of FPN-ResNet101 backbone of Mask R-CNN to jointly perform detection and tracking, without requiring additional training for tracking purpose.
The proposed single object tracker, TDIOT, applies an appearance similarity-based temporal matching for data association. In order to tackle tracking discontinuities, we incorporate a local search and matching module into the inference head layer that exploits SiamFC for short term tracking. 
Moreover, in order to improve robustness to scale changes, we introduce a scale adaptive region proposal network that enables  to search the target at an adaptively enlarged spatial neighborhood specified by the trace of the target.
 In order to meet long term tracking requirements, a low cost verification layer is incorporated into the inference architecture to monitor presence of the target based on its LBP histogram model.
Performance evaluation on videos from  VOT2016, VOT2018 and VOT-LT2018 datasets demonstrate that TDIOT achieves higher accuracy compared to the state-of-the-art short-term trackers while it provides comparable performance in long term tracking.

\end{abstract}

\IEEEpeerreviewmaketitle

\section{Introduction}

Visual object tracking (VOT) aims to locate a target object, specified at the first frame, in subsequent video frames. Despite significant progress in the latest methods, VOT still remains a difficult task due to many challenges such as fast motion, 
background clutters, partial occlusion, etc
\cite{sv4,sv5,ltBENC,mot2019,vot2020}. As a consequence of the high accuracy achieved by the deep object detectors,  
 tracking-by-detection (TBD) paradigm has become very popular among VOT methods \cite{mcpf, tbd2, l1dpfm,retinetrack,tracktor,tbdmot,DEEP_A}. There are two major approaches that are dominant within TBD paradigm. The first one converts a detector to a tracker by only inference stage customization \cite{icip2017, deepsort,tracktor,GURKAN}. The second one seeks to formulate detection and tracking as a joint optimization problem and applies offline retraining schemes for tracking purpose \cite{retinetrack,dandt,t-cnn}.
Although the former methods are often more computationally efficient, their overall performance is highly dependent on the detector baseline.
The latter methods can benefit more from the temporal information while detecting the target object.
However, joint optimization requires end-to-end training of the whole network which may not always be suitable due to computational burdens and lack of in-domain training data. 
Furthermore, despite their specialized training schemes, improvement in terms of tracking accuracy is still mediocre \cite{retinetrack}, \cite{tracktor}.

In this work, we adopt the first approach and introduce an inference architecture for tracking that enables to convert a deep object detector to an object tracker. 
Our main motivation is to design a single object tracking scheme which does not require additional training for tracking purpose and complex inference stage customization. As a result, the proposed tracker is suitable for real life scenarios where training is mostly impractical because of lack of available data or other constraints on complexity. 
Furthermore, in this paradigm, one may change the baseline detector with a recent one easily owing to modular architecture. 

We also address the data association problem which is present in almost all TBD paradigm trackers. This problem arises as a result of having multiple object proposals with the same class label at the object detector output.
In order to cope with this problem we use a cosine similarity based resemblance analysis to identify the target between all proposals. If no proposal is present, we employ SiamFC \cite{SiamFc} in short term and  Kernelized Correlation Filter (KCF) \cite{kcf} in long term  as an auxiliary tracker to estimate the location of the target object.

Moreover, we aim to grant target-awareness to the region proposal generation.
Conventionally in both one-stage \cite{yolo,ssd} and two-stage \cite{MaskRCNN,faster,UnitBox} object detectors, the candidate object proposals are generated based on the pre-defined anchors where their size and scale are fixed 
in offline training. The detection accuracy is improved by the two-stage object detectors in which a region proposal network (RPN) in the first stage generates candidate object bounding box proposals and in the second stage their class labels and final locations are estimated via classification and regression layers.
In our case, we use a two-stage deep detector, specifically Mask R-CNN \cite{MaskRCNN} because of its superior detection performance \cite{detectionsurvey}. However, in a real-world scenario, target dimensions frequently change due to target or camera motion and other factors such as partial occlusion. 
To grant flexibility to the RPN proposals, we present a novel scale-adaptive region proposal network (SRPN) which incorporates the conventional RPN with a target-aware proposal sampler and a hierarchical proposal selector. SRPN has two main benefits, it can generate proposals with any size based on the last tracked target's size and  can better localize the target object without additional training.  

The final problem that we focus on is the target object verification. In long-term tracking videos target object may disappear from the scene and reappear again after some time. To tackle with this problem, we include a target verification layer in the inference head, which employs Local Binary Patterns (LBP) based appearance representation to detect exit and entrance of objects. Moreover a local-to-global search strategy is integrated into the region proposal generation network to recover the target object after tracking discontinuities.

Key contributions of this work are summarized as follows:
\begin{itemize}

      \item We convert Mask R-CNN object detector to a video object tracker by placing  a novel inference architecture on top of its FPN-ResNet101 backbone.    The proposed tracker, Target-driven Inference for Deep Video Object Tracking (TDIOT), enables utilizing visual and temporal information of target object without additional training.
  
   \item We introduce a target-driven region proposal generation network which utilizes temporal properties of target object by using a proposal sampler. This allows proposals to adapt to variations on target size, scale and location throughout the video sequence.
   \item In order to detect object entrance and exit, we incorporate a target  verification layer into the inference architecture for long-term object tracking.
   The proposed layer executes a simple yet efficient target verification scheme that discriminates the LBP histogram based appearance model of the target by applying  local to global search.
  \item  We extensively evaluated TDIOT on a subset of commonly used VOT2016, VOT2018 and VOT-LT2018 datasets. Numerical results demonstrate that TDIOT has superior accuracy compared to the state-of-the-art short term trackers while it provides higher F1 score under viewpoint-change in long term tracking.
\end{itemize}

 \section{Related Work}

\textbf{Converting a Detector to a Tracker:} Recently CNN based object detectors have brought significant progress and clearly outperformed all other methods by effectively learning the spatial information via training \cite{MaskRCNN,faster,yolo,ssd,retina,FCOS,review_trac}. Since the state-of-the-art deep detectors are trained on still images, they cannot be effectively employed for video object tracking  without including the temporal information into the model. In the literature, several deep network architectures are proposed to take the advantage of temporal information in video processing.  In particular, multi-stream CNNs \cite{two1,multistream-J}, 3D ConvNets \cite{ Carreira2017,Xingjian2015} and recurrent networks \cite{rolo,lstm_m,st_lstm} are used for visual object tracking.  The main drawback of these models is they are far from being simple since the utilities used to capture the temporal information are complex and their performance highly depends on the model hyperparameters. 

Apart from  these approaches, TBD methods are becoming very popular since the deep object detectors provide very high accuracy on individual images. Two common approach under TBD paradigm is to convert a detector to a tracker either with the help of an inference architecture designed for tracking or by jointly training the detector and tracker. 
Motivation behind the conversion via the former approach is linking detection results across video frames with a low cost scheme, where the linking step is referred as data association.
 Under this framework, it is common to use a successful deep object detector like Faster R-CNN, Mask R-CNN, YOLO or SSD as the discriminative tool while Bayesian filtering or another simple state transformation method as the generative tool. In \cite{icip2017,GURKAN,erdem} the data association is accomplished with a late fusion scheme applied at the decision level.  In \cite{l1dpfm,deepass}, integration of the detector and tracker is achieved by a low level fusion formulation  to take advantage of both. 
Although very successful results are reported in these works design of  an appropriate fusion scheme constitutes the main difficulty. Moreover,
DeepSort \cite{deepsort}, as also one of the pioneers of the former approach, takes Faster R-CNN detection outputs and links them through the video frames by Kalman filtering. Data association is performed by Mahalanobis distance metric and the final tracked objects are estimated by appearance matching based on cosine similarity calculated on ReID feature maps. Although DeepSort employs a pre-trained Faster R-CNN detection model, a custom designed ReID network is trained for pedestrian tracking thus works for only one object class. Apart from other methods, Tracktor \cite{tracktor}  does not utilize any data association algorithm and exploits the regression head of Faster R-CNN object detector to regress the last tracked object bounding box (BB) to the new position. Tracktor++, extended version of \cite{tracktor},  employs two different motion models to transform detections before regression and incorporates a Siamese based ReID network trained for pedestrian tracking. TDIOT does  not exploit the head of a detector that limits the tracking capability with the regression efficiency of the deep detector. Instead,  we replace the head layer of a deep detector, in particular Mask R-CNN, with an inference architecture which enables utilizing spatio-temporal information and prevents the temporal discontinuities. 

In the literature a number of trackers are also designed under the second perspective that requires an end-to-end training for tracking. D\&T \cite{dandt} employs R-FCN \cite{rfcn} as the baseline detector and performs similarity learning based on a novel loss function. In inference stage,  the data association and tracking is accomplished by the class-wise linking score assigned to the candidate object patches. 
MaskTrack \cite{maskTrack} embeds a new tracking branch to Mask R-CNN to jointly perform the detection and tracking. In inference phase, data association is accomplished as a combination of three metrics quantifying object class, detection score and IoU. Similar to \cite{maskTrack} and \cite{dandt}, we filter the non-target candidate detections by class-wise filtering to ensure the semantic consistency throughout the video sequence.
RetinaTrack \cite{retinetrack} adapts a single stage detector, specifically RetinaNet \cite{retina}, to provide instance level feature maps. Aside from the conventional trackers which employ data association schemes, RetinaTrack learns to associate detections by adding instance level embeddings to RetinaNet’s post-FPN prediction sub-networks. In inference phase, detections are filtered by score thresholding and then surviving embedding vectors are compared to the previous tracking results via cosine distance and IoU metrics for estimating the final match for the target. Similar to \cite{retinetrack,deepsort}, we perform data association based on  appearance similarity which is particularly useful when the motion  uncertainty is high. 
However, we check the similarity in RGB space rather than on feature maps. 

\textbf{Efficient Proposal Generation:} Proposal generation is a crucial step in object detection, as it directly effects the tracking  performance. Former proposal extraction methods such as Selective Search \cite{ss}, EdgeBox \cite{edge}  are sensitive to appearance changes, partial occlusion etc. as well as they are computationally expensive. More recently, RPN is introduced as the proposal network of Faster R-CNN \cite{faster}. Conventional RPN is trained offline by using anchor boxes where scales and aspect ratios of the anchors are fixed during training. Because of the uniform anchoring, RPN has difficulties to deal with object candidates when large size and scale variations are present. To alleviate this drawback, \cite{garpn} proposes a guided anchoring region proposal network (GA-RPN) that allows generation of sparse and non-uniform anchors under the assumption of scale and aspect ratio of an object relates to its location. The proposed framework is optimized by an end-to-end training using a multi-task loss, including anchor localization and shape prediction losses. It is shown that GA-RPN improves overall mAP of  Faster R-CNN \cite{faster} and RetinaNet \cite{retina}. Apart from anchor based region proposal generator, recently, anchor free methods are also proposed \cite{FCOS,siamcar,Ocean,siambox} which usually have simple architectures and produce final object detections within a single stage. But despite the absence of anchors and proposal refinement module, which significantly reduces the computational load, their performance is limited especially on complex backgrounds.

Another drawback of vanilla RPN in tracking is the lack of target awareness. Since region proposals are generated for all detected object candidates, most of them are irrelevant to the target object. While using RPN in object tracking it is necessary to limit proposals to only those closely related to the target object to increase tracking performance by minimizing false detections \cite{pst,siamRPN,fewshot}. SiamRPN \cite{siamRPN} consists of Siamese sub-networks for feature extraction and proposal generation where their outputs are respectively fed into the classification and regression branch. Unlike RPN,  the correlation feature map of the target template is given as input to RPN of SiamRPN. While SiamRPN introduces a new RPN architecture, it still needs carefully designed anchor boxes. IDPF-RP \cite{GURKAN} introduces a novel Region Proposal Alignment (RPA) scheme to select object proposals related to target. RPA applies non-maxima suppression (NMS) on the proposals generated by Mask R-CNN and particle filter 
and highly improves the localization accuracy of tracking. Numerical results reported on VOT2016 dataset demonstrate that \cite{GURKAN} provides about 15\% higher success rate by using RPA scheme.
Inspiring from \cite{GURKAN}, in this work, we propose a Scale Adaptive Region Proposal Network (SRPN) where the target guided proposal sampler (TPS) and RPN of detector are integrated. Main idea behind the proposal generation scheme is forcing the agreement between two proposal subsets to ensure the best proposals are selected. TPS increases flexibility in size and scale of the proposal BBs by sampling the  proposals from the last tracked BB, and also imposes the target awareness by limiting the search region around the target object.

\textbf{Long-term Tracking :}  In order to fulfill requirements of a long term video object tracker, we extended TDIOT by including an object verification layer into the architecture.  Also a local-to-global search mechanism  is designed to enable detection of object entrance into the scene.
 While modifying  TDIOT for long term object tracking , our aim is to detect object entrance and exit independently from the target object class. In order to keep the computational load low, we avoid to use available triplet-loss based networks that can be trained for verification \cite{triplet,triplet_im}. We also seek to design a target verification layer such that being able to track general objects, not only specific objects (e.g. pedestrian, car etc.) as in \cite{tracktor,deepsort,deep_metric}.
We propose a simple yet efficient target verification scheme that discriminates the local binary patterns (LBP) \cite{lbp} histogram based appearance representations with  chi-square distance metric by applying local to global search for presence and absence detection of the target. Due to its discriminating capability LBP and its variants are previously used for several applications including object re-identification \cite{REid_survey}, \cite{lbp_3}. Differ from the existing forms, we monitor temporal changes of the LBP histogram representations to verify presence of the target while the target model is updated at object reappearance frames.

Similarly, existing long term trackers have the target verification, reappearance search and target model update mechanisms but in different forms.
DaSiamLT \cite{dasiam} which is the second best tracker in the VOT-LT2018 challenge, retrains SiamRPN \cite{siamRPN} with a distractor-aware training scheme that significantly improves dissimilarity learning. After a long offline training,  DaSiamLT detects target presence by a simple similarity score thresholding incorporated with a local-to-global search strategy. Differ from \cite{dasiam}, \cite{pattern_lt} uses SiamRPN as a candidate object BB proposal generator incorporated with a Siamese network designed as target presence-absence classifier. After a supervised classifier training, the candidate BB having the highest classification score is decided as the tracked object. When non of the proposals is classified as the target, the tracker attempts to re-detect the target via a global search followed by reclassification.
\cite{MBMD}, the winner of VOT-LT2018 challenge, introduces a long term tracker that includes a Siamese network based regression layer. As a verification network, it uses MDNet \cite{mdnet}, that takes candidates generated by RPN as its input and outputs their probability of being a foreground or a background object. The final confidence score that indicates whether the object is present or absent is estimated with the scores taken from both regression and verification modules. After disappearing, the tracker switches from local to global search for re-detection, where the global search is conducted by conventional sliding window search. Apart from \cite{pattern_lt} and \cite{MBMD} which perform online updating on the verification network to adapt the appearance changes of the tracked object, our verification layer does not require any optimization or re-training during tracking.

\begin{figure*}[!t]
\centering
\includegraphics[width=0.9\textwidth]{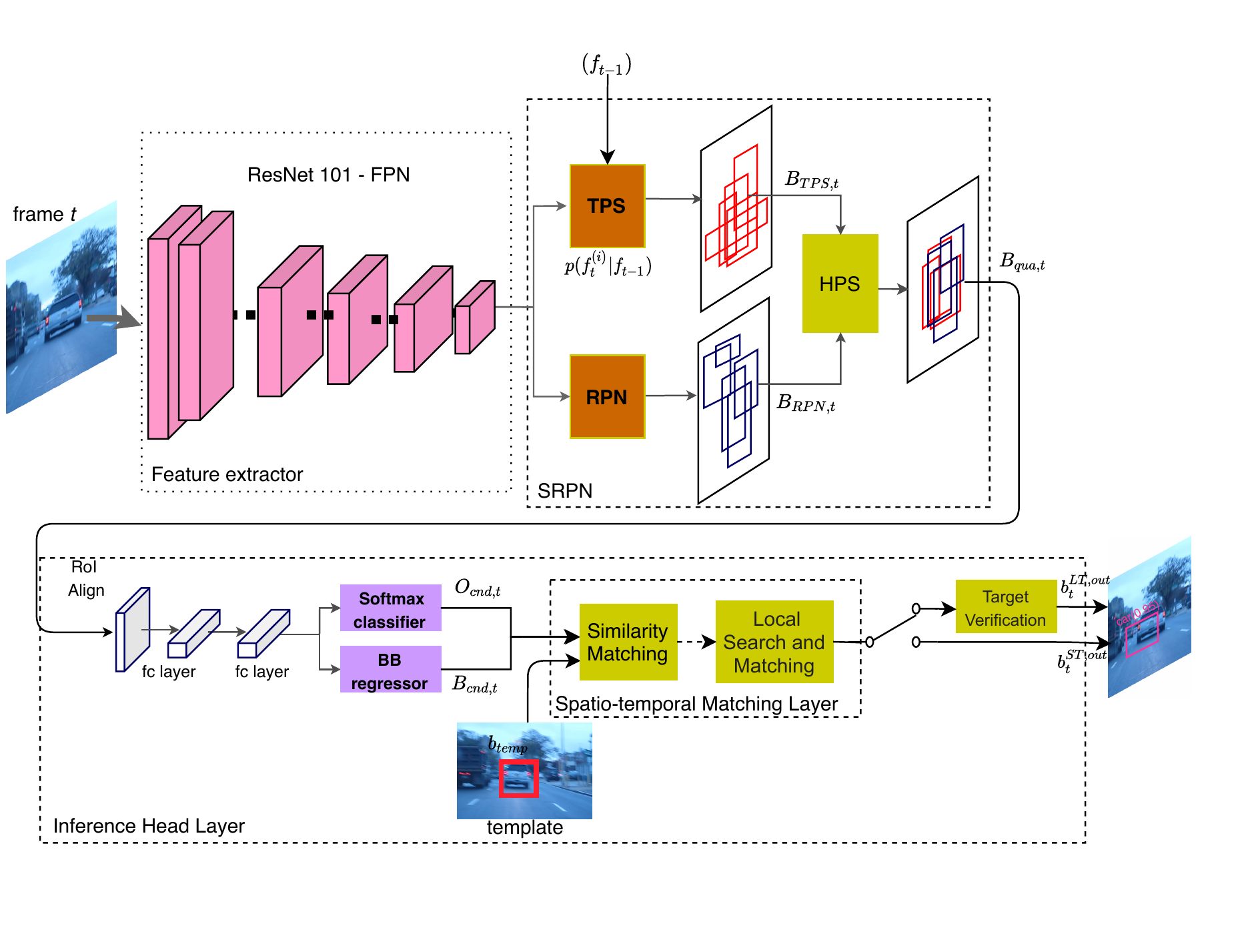}
\vspace{-0.6cm}
\caption{Inference architecture of TDIOT.} 
\label{fig_tamask_architecture}
\end{figure*}

\section{Inference Architecture}
\label{section_mask_tracker}

In order to tackle both data association and  temporal alignment problems, we propose to replace the training architecture of a deep object detector, specifically Mask R-CNN, by a novel architecture at inference stage. In particular, we replace the region proposal network as well as the head layers of Mask R-CNN with the proposed inference layers while we use the backbone of vanilla Mask R-CNN as it is. The proposed video object tracker, TDIOT, does not require additional training for tracking and makes use of the trained object detector.

The proposed inference architecture, given in Fig. \ref{fig_tamask_architecture}, consists of three main blocks; a ResNet 101 based feature pyramid network (FPN) backbone that provides feature map for the input video frame, Scale Adaptive Region Proposal Network (SRPN) that generates object bounding box proposals by taking feedback from the head, and an inference head layer that outputs the tracked target object bounding box along with its object class label. The inference head layer includes a similarity matching layer, local search and matching layer. 
In order to detect the exit and entrance of objects that may arise from occlusion or being out-of-view, a target verification layer is also included into the inference head layer for long term tracking. 
The proposed tracker, TDIOT, employs the same backbone architecture with vanilla Mask R-CNN \cite{MaskRCNN}. In the following, we present SRPN and the inference head layer.
\subsection{Scale Adaptive Region Proposal Network}
\label{section_mrpn}
At the inference stage, RPN of vanilla Mask R-CNN detects a set of candidate object BB proposals.
Feature maps corresponding to the candidate BBs are cropped from the features extracted by Resnet101-FPN backbone network \cite{resnet}. Candidate object proposal BBs along with the corresponding feature maps are fed into the binary classification and regression layers that output the aligned object proposal BBs along with the confidence scores. In our notation $B_{RPN,t}=\{b_{RPN,t}^{j} \}_{j=1}^{R}$ denotes the set of proposal BBs for frame $t$ where the number of proposals is fixed to $R$.

Conventionally  RPN employs a set of anchors where their size and scales are fixed after offline training of the object detector.  In order to effectively deal with the scale variations of the tracked object and to employ the temporal information provided by the video data, we replace RPN of the vanilla Mask R-CNN with Scale Adaptive Region Proposal Network (SRPN). SRPN incorporates Target-guided Proposal Sampler (TPS) to RPN of vanilla Mask R-CNN. Outputs of TPS and RPN are fused at a Hierarchical Proposal Selection (HPS) layer that outputs a set of candidate object proposals. In the following, we formulate TPS and HPS layers in detail.

{\bf Target-guided Proposal Sampler }: The learned anchors may not perfectly fit the size of objects present in an unseen scene thus might lead to misdetections. 
In order to alleviate this problem, Target-guided Proposal Sampler (TPS) that provides additional $P$  proposals  $B_{TPS,t}=\{b_{TPS,t}^{i} \}_{i=1}^{P}$ at frame $t$ is included into the inference architecture.

Inspired by the particle samplers of Bayesian filters \cite{bayes3}, TPS is realized by a Gaussian sampler with the transition model formulated by Eq. \ref{eq_sampling}.

\begin{equation}
\label{eq_sampling}
p(f_t^i|f_{t-1})\sim \mathcal{N}({\mathbf{\mu}}_{t},{\mathbf{C}}_{t})  , \hspace{0.2in} i=1,...,P.
\end{equation}

 In Eq. \ref{eq_sampling}, the vector ${{f}}_{t-1}=(x_{t-1}, y_{t-1},w_{t-1},h_{t-1} )^{T}$  and ${{f}}_{t}^{i}=( x_{t}^{i}, y_{t}^{i},w_{t}^{i},h_{t}^{i} )^{T}$ respectively represent the last tracked object BB, $b_{t-1}$, and the $i$th sampled proposal BB, $b_{TPS,t}^{i}$; where ($x$, $y$) is the center pixel coordinates of BB, and $w$ and $h$ respectively denote its width and height of it. 
 
 The idea behind inserting TPS into the proposal network is to effectively tackle size variations of the tracked object. Hence, as it is formulated in Eq.\ref{eq_sampling}, TPS proposals are sampled independently via a Gaussian sampler $ \mathcal{N}( {\mathbf{\mu}}_{t},{\mathbf{C}}_{t}) $ with the statistics  $ {\mathbf{\mu}}_{t}= {{f}}_{t-1}$ and  ${\mathbf{C}}_{t}=\text{diag}(\sigma^2_{x_{t}},\sigma^2_{y_{t}},\sigma^2_{w_{t}},\sigma^2_{h_{t}})$. Specifically, by setting the mean of the sampler for frame $t$ to ${{f}}_{t-1}$, the state of the object tracked at $t-1$,  we aim to improve localization accuracy and scale adaptation of the tracking.
In particular, the first two attributes of the sampler mean vector, $ (x_{t-1}, y_{t-1})$, determine where spatially to look for the target object and the last two, $(w_{t-1}, h_{t-1})$, determine its expected BB size. 

Furthermore, the sampler covariance matrix $\textbf{C}_{t}$ enables us to adaptively adjust the search region as well as the anchor size. In particular, $\sigma^2_{x_{\scriptsize{t}}}$ and $\sigma^2_{y_{\scriptsize{t}}}$ model the expected horizontal and/or vertical translation of the target object between two consecutive video frames. 
By increasing these variances, we enlarge the search region around the center of the latest tracked target BB. This also provides more flexibility to the tracker in terms of adopting to a high motion object. 
On the other hand, $\sigma^2_{w_{\scriptsize{t}}}$ and $\sigma^2_{h_{\scriptsize{t}}}$ model the expected size changes of the tracked object BB that allows searching by unfixed anchor sizes hence improves robustness to scale changes. 
 
{\bf Hierarchical Proposal Selection}: For each frame at time $t$, proposals generated by the RPN, $B_{\text{RPN},t}=\{b_{\text{RPN},t}^{j} \}_{j=1}^{R}$, and TPS, $B_{\text{TPS},t}=\{b_{\text{TPS},t}^{i} \}_{i=1}^{P}$, are passed to HPS that filters the unqualified proposal BBs based on intersection-over-union criteria given in Eq. \ref{eq_iou}. 
\begin{equation}
\label{eq_iou}
    \textnormal{IoU}(i,j)=\frac{|b_{\text{TPS},t}^{i} \cap b_{\text{RPN},t}^{j}|}{|b_{\text{TPS},t}^{i} \cup b_{\text{RPN},t}^{j}|}, i=1,..,P, j=1,..,R.
\end{equation}

Specifically, HPS layer calculates IoU scores for all pairs from RPN and TPS proposals and then selects pairs above a threshold $\tau_{k}$ to obtain a qualified output BB set $B_{k,t}$ (Eq.\ref{eq_fusing}). Here $k$ denotes the level of the threshold in a hierarchical thresholding scheme.

\begin{equation}
\label{eq_fusing}
    B_{k,t} = \{B_{\text{TPS},t} \cup B_{\text{RPN},t}\; |\;  \textnormal{IoU}(i,j) > \tau_{k} \}  
\end{equation}

The reason that we propose a hierarchical thresholding scheme rather than a simple thresholding approach is to limit the candidate BBs to only those we have high confidence. This not only allows better localization performance, it also reduces overall run time by significantly decreasing the number of proposals passed on to the head layer. Therefore, we start with a higher IoU threshold $\tau_{max}$, which ensures higher overlap ratio, i.e. consensus, between proposals. If there are no candidates to satisfy the condition, so that the set of qualified BBs at level $k$, $B_{k,t}$ is empty, then we decrease the threshold and search for a new qualified BB set. If $ B_{k,t}$  remains empty for a predefined minimum threshold level $\tau_{min}$, than we conclude that there's no agreement between TPS and RPN for that frame. The reason that we stop the search at a predefined threshold lower bound is that when the threshold drops below a certain point the matching pairs do not relate well with each other since their consensus is too low. This may arise from either when TPS drifts from the target or when RPN fails to detect object BBs. 
Final qualified BBs after the scheme are collected in the set $B_{\text{qua},t}=\{b_{\text{qua},t}^{i}\}_{i=1}^{H}$. Algorithm \ref{alg1} explains how HPS works.

\begin{figure*}
\centering
\subfloat{\includegraphics[width=0.9\linewidth]{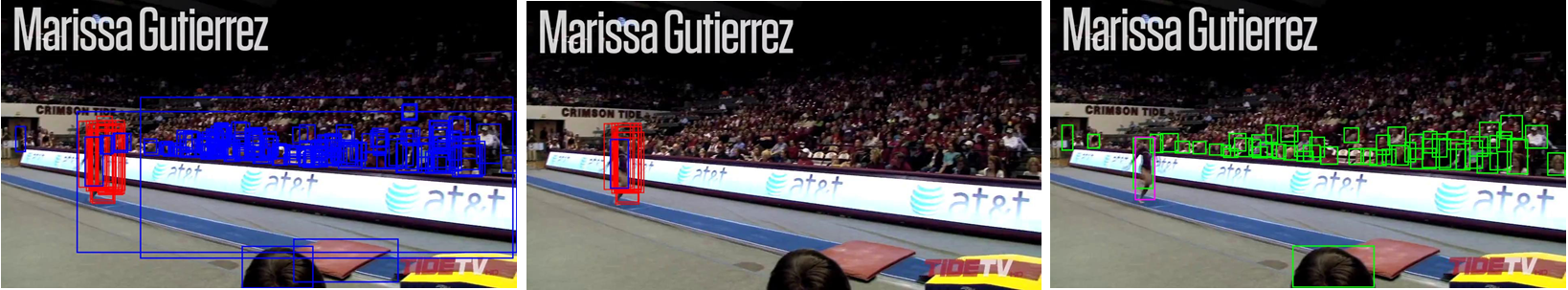}}
\caption{ Gymnastics3 video sequence, frame no 5. (left) Proposals generated by RPN (blue BBs) and TPS (red BBs). (middle) Qualified proposals selected by HPS. (right) Tracked object BBs provided by the head layer of Mask R-CNN (green) and TDIOT (magenta).}
 \label{stdeffectbg}
\end{figure*}

\begin{figure}[t]
\centering
\includegraphics[width=3.5 in]{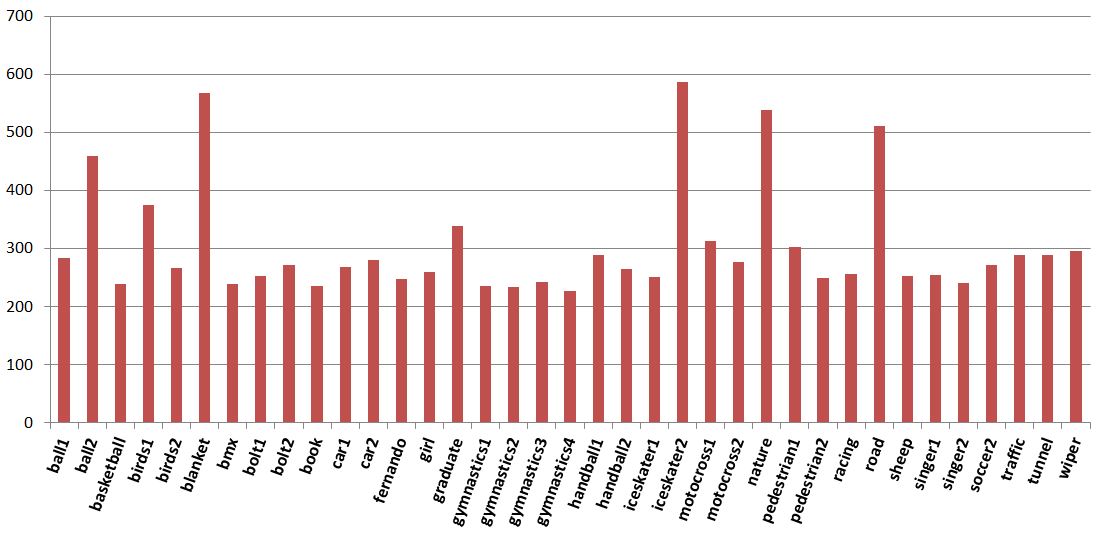}
\caption{ Number of proposals passed from SRPN to the head layer in VOT2016 video sequences.  RPN of vanilla Mask R-CNN passes 1000 proposals to the head.}
\label{uyumlu}
\end{figure}

Fig. \ref{stdeffectbg} demonstrates the impact of the introduced SRPN.
We observe that the qualified BBs are clustered around the target object while many of the redundant BBs are accurately filtered that yields a well localized tracked object in contrast to the object detected by Mask R-CNN. It is also notable that the number of qualified candidate BBs is adaptively changed from frame to frame but significantly reduced  compared to the fixed number of BBs generated by vanilla RPN. 
  Fig. \ref{uyumlu} reports the number of qualified proposals in $B_{qua,t}$ obtained by TDIOT on VOT2016 dataset.
  Specifically, for each frame, RPN of vanilla Mask R-CNN passes $R=1000$ proposals to the head, however in our performance tests the number is reduced to 300 proposals on average that means decreased by 70\% where  TPS samples 
$P=200$ proposals.

\begin{algorithm}[ht]
\DontPrintSemicolon
  
  \KwInput{$ B_{\text{TPS},t}$ and $ B_{\text{RPN},t}$ 
 \\Set IoU thresholds in descending order $\{\tau_k\}^m_{k=1}$}
  \KwOutput{$ B_{\text{qua},t}$}
  k=1 \\
  \While{$k \leq m$}
  {    $B_{k,t} = \{B_{\text{TPS},t} \cup B_{\text{RPN},t}\; |\;  \textnormal{IoU}(i,j) > \tau_{k} \} $

 \If{$B_{k,t} \neq \phi$}
       { $B_{\text{qua},t}=B_{k,t} $
       \\ \textbf{break}}
       k++       } 
\If{$B_{k,t} =\phi$}
 {$B_{\text{qua},t}=B_{\text{RPN+TPS},t}$}
\caption{HPS algorithm}
\label{alg1}
\end{algorithm}


\subsection{Inference Head Layer}
\label{section_feedback}

 As in vanilla Mask R-CNN head,  Inference Head Layer first applies a localization alignment on the qualified candidate BBs passed from SRPN. In particular, the feature maps corresponding to the candidate BBs are resized by RoI align and passed into the regression layer that provides localized BBs while the classifier layer assigns the object class labels of BBs along with the objectness scores. We introduce Spatio-temporal Matching Layer (SML) that constitutes the core module of our tracker. Specifically, in order to suppress non-target detections,  Similarity Matching (SM) module included in  SML performs data association  between the consecutive frames. Moreover, a Local Search and Matching (LSM)  module is added to  the pipeline in case of a miss detection  (Fig. \ref{fig_tamask_architecture}). 
 Following paragraphs give the details of SM and LSM modules.
 
\subsubsection{\bf Similarity matching}

 The set of qualified proposal BBs, $B_{\text{qua},t}$, is fed into  the head layer. The BB regressor of the inference head layer outputs all candidate object BBs while their class labels and objectness scores are assigned by softmax.  
Let, $B_{\text{cnd},t}=\{b_{\text{cnd},t}^{i}\}_{i=1}^{D}$ and $O_{\text{cnd},t}=\{o_{\text{cnd},t}^{i}\}_{i=1}^{D}, D<H$ denote the set of candidate object BBs and their class labels, respectively. 
Among the candidates, SM module estimates the final tracked object BB based on appearance similarity. 
In order to reduce false alarms, SM first eliminates the candidates having object class label different from the target. 
Hence the register $d_t(i), i=1,..,D,$ turns out to $1$ when the class label of the target object $o_{\text{gt},t}$ and a candidate object $o^{i}_{\text{cnd},t}$ are the same.
 \begin{equation}
 d_t(i)=\vmathbb{1} [o^{i}_{\text{cnd},t}=o_{\text{gt},t}]
	\label{Eq:ChiSquare2}
\end{equation}

SM estimates the final tracked object among the filtered candidates based on appearance similarity. 
As in many other recent trackers \cite{deepsort,retinetrack}, we use cosine similarity metric for this purpose. 
To take into account  the scale and the appearance changes, we define the cosine similarity score as in Eq. \ref{cos}.  $c_t(i)$ quantifies similarity between the template patch, $p_{\text{temp}}$, and $p_{\text{cnd},t}^{(i)} $, the patch corresponding to the i${\it th}$  candidate BB of the class-wise filtered candidate set, where $f(.)$ defines an embedding function that maps the RGB image to Fourier domain.

\begin{equation}
\label{cos}
    c_t(i)=  \frac{f^T(p_{\text{temp}}).f({p_{\text{cnd},t}^{i}})}{\|f(p_{\text{temp}})\|_2.\|f({p_{\text{cnd},t}^{i}})\|_2} 
\end{equation} 

SM outputs the candidate BB having the highest score as the final tracked object patch, $p_t^{out}$, since the similarity score indicates the degree of confidence to where the target object is located. 

\begin{equation}
\label{asm}
    p_t^{\text{out}}=\argmax_{p_{\text{cnd},t}^{i}} (c_t(i))
\end{equation} 

Accordingly the target template,  $p_{\text{temp}}$, is updated by the latest tracked object patch, $ p_t^{\text{out}}$, to adapt to the target appearance changes.  
 $b_t^{\text{out}}$ is set to the BB of the tracked object patch, $p_t^{\text{out}}$.

Similarity matching module enables to keep tracking the target with low cost especially under smooth motion.
However when the deep detector fails to detect the target (because of blur, busy background, etc.), the tracker also fails that may result in drifts from the target object. To alleviate this drawback we integrate Local Search and Matching (LSM) module into the inference head layer.

\subsubsection{\bf Local Search and Matching}
Our LSM module borrows the matching strategy of Kernelized Correlation Filter (KCF) \cite{kcf} as an effective low cost solution. It is also shown that replacing KCF by SiamFC \cite{SiamFc} improves performance for short term tracking. 

In particular KCF \cite{kcf} is a Kernel Ridge Regression method trained with many sample candidate patches around the object at different translations.  In our tracker, the candidate locations are defined as shifted variants of the last tracked target location.
LSM module of TDIOT employs this learned filter to determine the best candidate object patch, $ p_t^{\text{out}}$, based on the detection score which is also called peak-to-side lobe ratio (PSR).  Higher PSR value indicates the degree of confidence to which the target object is exactly located at that location. Differ from conventional KCF, where the object template is updated during tracking by weighted average of the previously estimated tracked object, in our tracker, the target model is updated by the last tracked object patch.

As an alternative to KCF, we have also evaluated  SiamFc \cite{SiamFc} that performs similarity learning by using Siamese networks. Basicly,  we use SiamFC as a matching tool  not a tracker, where \textit{the matching} refers to determine the best candidate within a search region. More specifically, our LSM module takes  a set of candidate  object locations proposed by SiamFC and estimates the final tracked object as the one having the maximum similarity with the target model. SiamFC generates the candidate object proposals by a sliding window in a local search area which is adjusted to approximately four times of the last tracked target patch.
Different from \cite{SiamFc} where the object template is fixed, we update the target model, $p_{\text{temp}}$, during tracking by the last tracked object patch. We empirically show that SiamFC increases accuracy for short term tracking.
This is because the similarity learning improves robustness to appearance changes hence reduces false matches. 

Although we significantly reduce the miss detection ratio we still need to extend the model to tackle object entrances and exits from the scene. Next section presents extension of TDIOT to long-term tracking.

\section{TDIOT-LT : TDIOT for Long-term Tracking}
\label{subsec:LT}
Long term object tracking has its own demands such as robustness to object entrance and exits, occlusion, abrupt appearance changes and viewpoint change. Thus, employing a tracker for long term tracking requires care. 
For this reason, we insert extra layers to the inference head of TDIOT to meet requirements of the long-term tracking. 
More specifically, a target verification (TV) layer is implemented with the aim of confirming presence of the target object. Furthermore we introduce a local-to-global search strategy to adaptively enlarge the search region at tracking discontinuity frames.

As a low cost solution we propose a LBP \cite{lbp} based detection scheme that allows to verify presence of the target object at each frame.
Specifically, we monitor the tracking from beginning to the end and confirm the presence / absence of the tracked object at each frame $t$ by using a verification score $v_t$, 
 \begin{equation}
  v_t = \chi^2_{t} - \dfrac{1}{t-1} \sum_{l=1}^{t-1} \chi^2_{l} 
	\label{Eq:Score}
\end{equation}
where $\chi^2_{t} $ is the chi-square distance calculated at frame $t$ by Eq.\ref{Eq:ChiSquare}.
 
In particular, we model the target template by its {\em N} bin LBP histogram, $h_{\text{temp},t}(.)$,  and update the target model at each frame $t$ based on the tracked object patch. Similarly,  $h_{t}^{\text{out}}(.)$, {\em N} bin LBP histogram of the object patch, $b_{t}^{\text{out}}$,  declared as the final tracked object at frame $t$ is calculated. Similarity between these histograms is measured by the chi-square distance given in Eq.\ref{Eq:ChiSquare}.

 \begin{equation}
   \chi^2_{t} =  \sum_{i=1}^{N}\dfrac{(h_{\text{temp},t}(i) - h_{t}^{\text{out}}(i))^2}{(h_{\text{temp},t}(i) + h_{t}^{\text{out}}(i))}.
	\label{Eq:ChiSquare}
\end{equation}

It is clear that chi-square distance calculated by  Eq.\ref{Eq:ChiSquare} does not rapidly change since the object exits the scene frame-by-frame, same as when it is occluded by another object. To tackle these smooth appearance changes we monitor chi-square distance variations from the mean chi-square distance of previous frames and formulate the verification score $v_t$ by Eq.\ref{Eq:Score}. In Eq.\ref{Eq:Score}, the mean chi-square distance of the last $(t-1)$ frames is calculated where $l=1$ corresponds to the first frame that the target object appears. In other words, we reset the mean chi-square distance calculation at each appearance of the target. The tracker confirms the target disappearance by simply thresholding the verification score $v_t$. Even though we calculate the chi-square distance for each frame, in practice, in order to minimize the computational load, the verification stage is activated for only the outputs of LSM layer, since the tracker keeps tracking the target without and discontinuity for the rest of the time. 

\begin{figure}
\centering
\subfloat{\includegraphics[width=1\linewidth]{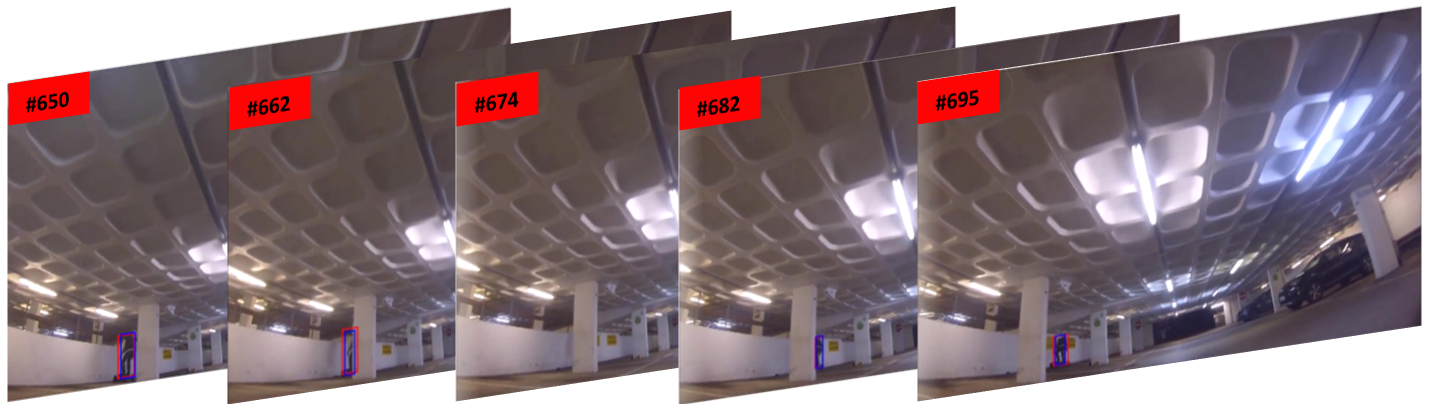}}

\subfloat{\includegraphics[width=1.01\linewidth]{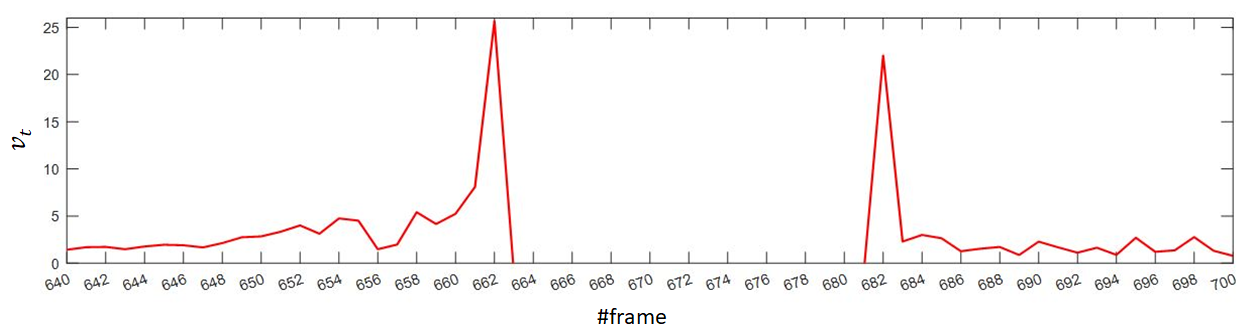}}
\caption{The tracking results of  longboard video sequence from VOT-LT challenge. First row: ground truth BB (blue BB). TDIOT-LT (red BB). Second row: verification scores.
 }
\label{lbpeffect}
\end{figure}

In order to detect entrance of the object into the scene, we modify the search strategy of the proposal sampler SRPN in such a way that allows to search the object within an adaptively enlarged search region. This is achieved by iteratively increasing the variance parameter of the state transition model formulated by Eq.\ref{eq_sampling}, after disappearance of the object. This is achieved by repeatedly passing the candidate proposal set to SM and LSM layers for enlarged search regions controlled by the increased variance parameter. In our work, simply we double the variances which model the horizontal and vertical translation. Fig. \ref{lbpeffect} illustrates how the verification score varies through the video frames. 
We observe that TV layer detects the exit and entrance frames for the target with a small delay.
In particular the object leaves the scene at frame 663 and re-enters at frame 682 and these frames are respectively detected as frame $663$ and frame $683$ by the tracker.

\section{Performance Evaluation}

In this section, we evaluate performance of the proposed tracker, TDIOT, on various scenarios. 
First we formulate the performance metrics, and then present the ablation study in which we examine the impact of each individual layer of the proposed tracker. 
Finally, we compare TDIOT with the other state-of-the-art trackers on three benchmark datasets: VOT2016 \cite{VOT2016}, VOT2018 \cite{VOT2018} and VOT-LT2018 \cite{VOT2018}.

\textbf{Hyperparameters for tracking:}
At the inference phase, we keep all of the hyperparameters used by vanilla Mask R-CNN except the objectness score threshold. 
The original threshold of objectness score $0.7$ is replaced by $0.2$ 
to allow most of the detected object BBs remain to be evaluated in further steps of the inference head.In TPS layer, proposals are sampled from a Gaussian distribution where the covariance matrix ${\mathbf{C}}_t$ in Eq.\ref{eq_sampling} is selected as $\text{diag}(5,5,5,5)$. The number of proposals sampled by TPS is
$P=200$, while RPN produces $R=1000$ proposals. For the hierarchical proposal selection presented in subsection \ref{section_mrpn}, we employed a three level selection procedure where thresholds are set to $\{\tau_{k}\}_{k=1,2,3} =\{0.8,0.5,0.3\}$, respectively. 

\subsection{Datasets and Evaluation Metrics}

We focus on two benchmarking datasets, VOT2016 and VOT2018, to evaluate short-term (ST) tracking performance of TDIOT. We used a Mask-RCNN model pre-trained on COCO benchmark, so although each dataset consists of 60 fully annotated videos, we only used videos in which target object classes are defined in COCO benchmark. So the actual numbers of videos used in our experiments are 37 for VOT2016 and 31 for VOT2018.

ST tracking performance is reported by using common evaluation metrics such as success plot, accuracy and robustness. More specifically, success plot is the distribution of success rate versus Intersection-over-Union (IoU) threshold where the success rate is the ratio of the successful frames whose overlap is larger than the given threshold. Accuracy (A) is formulated as the average $\textnormal{IoU}$ calculated over all successfully tracked video frames and robustness (R) measures how many times the tracker drifts off the target. We also define a new metric called tracking continuity (TC) to analyze impact of the proposed target-driven tracking scheme. TC measures how long can TDIOT track the target object without any interruption. It normalized by the length of the video sequence, hence TC is equal to 1 when the tracker successfully tracks the target object in all frames.

To evaluate the performance for Long-Term (LT) tracking, we   report tracking results on VOT-LT2018 benchmark which includes 35 challenging sequences of various objects. 
In LT tracking, sequences are much longer, and the target may leave the field of view for longer durations. 
On average, VOT-LT2018 dataset contains 4300 frames and 12 target disappearances. Apart from ST tracking, LT trackers need to output a prediction certainty score in each frame where target with the score below the threshold $\tau_\theta$ is assumed to left the frame. So common performance metrics, such as accuracy or success rate, may not be effective. In this paper, we use three LT tracking performance measures used in \cite{VOT2018}: precision (Pr), recall (Re) and their harmonic mean; F-score (Eq.\ref{fmeasue}).

\begin{equation}
\label{fmeasue}
 F(\tau_\theta)=2Pr(\tau_\theta)Re(\tau_\theta)/(Pr(\tau_\theta)+Re(\tau_\theta))
\end{equation}

As can be seen from Eq.\ref{fmeasue}, each measure directly depends on prediction certainty threshold $\tau_\theta$.
In the end, final value for each metric is obtained by selecting specific $\tau_\theta$ value which maximizes the F-score. This implies that each tracker can reach their highest performance at different threshold values. In our tracker, we use zero-one thresholding in which 1 indicates presence of the target object in the frame whereas 0 indicates its absence.

\subsection{Ablation Study}

In order to evaluate impact of the proposed layers, we analyzed performance of TDIOT on VOT2016 dataset for various scenarios which focus on different aspects of the system. 
Firstly, to show the effect of SRPN layer, we consider three different scenarios where the proposals passed out to the inference head layer are changed. 
The first scenario includes original TDIOT where the proposal set includes qualified proposals that are selected by SRPN ($B_{\text{qua}}$).
In the second one, we pass all proposals generated by both RPN and TPS to the head layer ($B_{\text{RPN+TPS}}$) without elimination.
In the last scenario only vanilla RPN proposals are passed out to the head layer ($B_{\text{RPN}}$). 
Table \ref{mrpnatt} compares the performances achieved by the trackers which use these three proposal sets. 
The notation KCF/Siam denotes the tracker utilized in the \textit{Local Search and Matching Layer}.

\begin{table}[ht]
\caption{ The Impact of SRPN on tracking performance. Accuracy (A), Robustness (R) and Inference time per frame (IT) of different trackers.  $\uparrow$ - the higher the better and $\downarrow$ - the lower the better.} 
\begin{center}
  \begin{tabular}{  l | l |c c c c }
    \hline  
	 Tracker & Proposal Set &A$\uparrow$& R$\downarrow$&IT (ms)$\downarrow$ \\ \hline \hline
 
   	&$B_{\text{qua}}$& {0.620} & {0.226}	&{649}	\\ 
  TDIOT-KCF  &$B_{\text{RPN+TPS}}$&0.547	&0.470	& 2170	\\  
 &$B_{\text{RPN}}$	& 0.544	& 0.460&	1920	\\ 
 \hline \hline
 
   &$B_{\text{qua}}$& {0.620}	&  {0.198}	& {617}\\ 
  TDIOT-Siam &$B_{\text{RPN+TPS}}$&0.583	& 0.438	&2270	\\  
	&$B_{\text{RPN}}$&  0.579	& 0.437&	2220	\\ 
 \hline\hline
 
   Mask R-CNN	&$B_{\text{RPN}}$& 0.602	& 0.580	&	344\\ 
   \hline
  \end{tabular}
  \label{mrpnatt}
  \end{center}
\end {table}

As can be seen from Table \ref{mrpnatt}, TDIOT-Siam and TDIOT-KCF achieve best  performance with the qualified proposal set. This shows that proposed SRPN layer has a significant impact on overall tracking performance. 
In particular, 24\% improvement in robustness indicates that elimination of the unqualified proposals prevents the tracker to drift away from the target object, while improvement in accuracy shows that the highly qualified proposals can lead to well localized BBs. 
In Table \ref{mrpnatt}, we also report the inference time (IT) to analyze the impact of proposal quantity on speed.
Tests run  on a PC with with Intel Core i7 4790 CPU 3.6 GHz and GeForce GTX TITAN X GPU. 
It is important to note that, SRPN significantly decreases IT, since the number of proposals in each frame are respectively 1200 and 1000 in $B_{\text{RPN+TPS}}$ and $B_{\text{RPN}}$ set, while it is 300 on average in the proposed method. 
We also compared our proposed tracker with the vanilla Mask R-CNN and report that the accuracy increases about 2\%, while robustness decreases 38\% and 35\% for TDIOT-Siam and TDIOT-KCF, respectively. 
This clearly demonstrates the improvement gained by including the spatio-temporal information into the visual object tracking.

In order to demonstrate the improvement achieved by the Similarity Matching layer, we design two variants of our tracker, which are denoted as “noCF”, and “HObj” respectively. 
“noCF” is the case where class information is not considered in similarity matching layer and “HObj” is the case where the objectness score is used instead of correlation similarity. 
In case of ``noCF'', where candidate detections with the false object class label are not eliminated, correlation metric may not be able to specify the correct object BB, because of higher false detection rate. 
As a result of this, it can be seen in Table \ref{AB5} that ``noCF'' case has increased robustness about 10\% and 7\% in TDIOT-KCF and TDIOT-Siam, respectively. 
In the other test case (``HObj''), choosing the BB with the highest objectness score causes tracking instabilities, hence it mostly results in selecting the wrong target BB, hence robustness increases about 4\% and 1\% in TDIOT-KCF and TDIOT-Siam, respectively.
We also report the TC metric that measures the tracking capability of the tracker without  any interrupt.
It can be concluded that the proposed TDIOT-Siam provides the highest TC with the help of Spatio-temporal Matching layer which is specifically designed to improve temporal connectivity. Table \ref{AB5} also reports the performance for the baselines.
As expected, R and TC of Mask R-CNN are not satisfactory because of its image-wise processing scheme that yields about $20\%$ miss detection rate in video object tracking.
Also stand alone performance of SiamFC makes clear the need for inclusion of the temporal information into the tracking model.  

\begin{table}[ht!]
\caption{Effects of Similarity Matching layer for different scenarios. TC indicates tracking continuity metric. The meanings of A, R, $\uparrow$ and $\downarrow$ are the same with the Table \ref{mrpnatt}.}
\begin{center}
  \begin{tabular}{ l  c  c c }
    \hline  
{Tracker}  & A$\uparrow$	& R$\downarrow$ & TC$\uparrow$\\  \hline \hline
{TDIOT-KCF} 	&{0.620} &{ 0.226}&{0.613}

\\ 
  TDIOT-KCF-noCF	&0.600 & 0.323&0.550

\\ 
 TDIOT-KCF-HObj  	&0.630 &0.271&	0.563
\\
 TDIOT-Siam   	&0.620 &	{0.198}&{0.656}
\\	
 TDIOT-Siam-noCF  	&0.610 &	0.289&0.584
\\	
 TDIOT-Siam-HObj 	& 0.630&	0.210&0.660
\\

  {Mask R-CNN} &0.602	&0.580 &0.168\\ 
 {SiamFC} &0.490	&0.330&0.519\\ 
 \hline 
  \end{tabular}
  \label{AB5}
  \end{center}
\end {table}

To evaluate the contribution of LSM layer, we conducted an analysis by comparing the performance of proposed tracker with the alternate matching methods. 
In Table \ref{lsma}, ``TDIOT-KCF'' and ``TDIOT-Siam'' denote the proposed tracker variants with KCF and SiamFC as the local search and matching strategies and ``TDIOT'' indicates the case where the LSM layer is completely left out. 
As expected, robustness of TDIOT is significantly higher (about 20\%) compared to the TDIOT-KCF/Siam, since it does not have any mechanism to handle the discontinuities during tracking. Also even though TDIOT-KCF and TDIOT-Siam has the same accuracy, the latter has better robustness since CNN based similarity matching scheme provides better discriminative capability.

\begin{table}[ht!]
\caption{Effects of local search and matching layer. The meanings of A, R, $\uparrow$ and $\downarrow$ are the same with the Table \ref{mrpnatt}.}
\begin{center}
  \begin{tabular}{ l  c  c c c}
    \hline  
{Tracker}  & A$\uparrow$	& R$\downarrow$  \\  \hline \hline
 TDIOT 	&0.630 &	0.407
\\
{TDIOT-KCF}  	&{0.620} &{0.226}
\\ 
 TDIOT-Siam   	&{0.620} &{0.198}
\\	
 \hline 
  \end{tabular}
  \label{lsma}
  \end{center}
\end {table}
  \vspace*{-\baselineskip}
\subsection{Comparison with State-of-the-Art Short-term Trackers}

\begin{figure*}[ht]
    \centering
    \begin{subfigure}[ht]{2.4 in}
        \includegraphics[width=\linewidth]{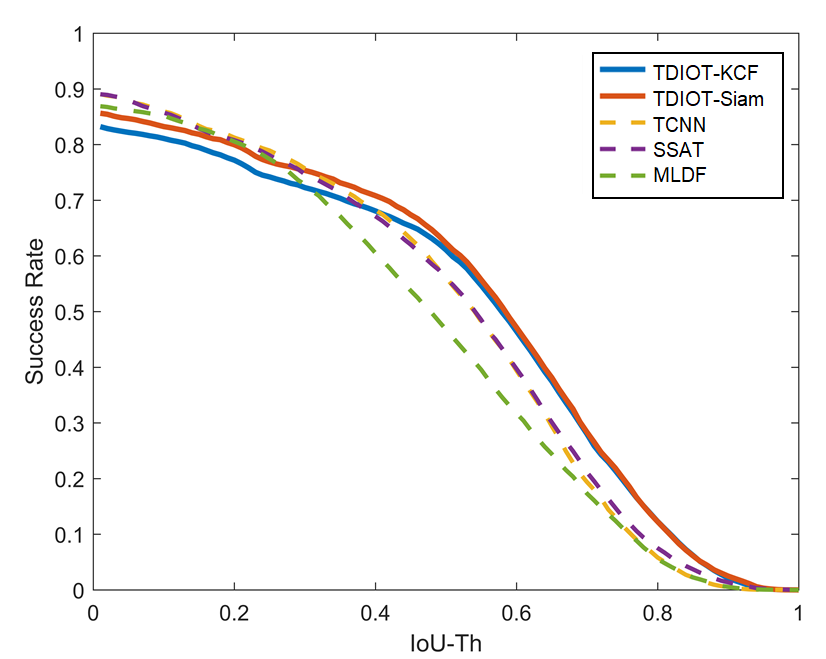}
        \caption{}
    \end{subfigure}
    \begin{subfigure}[ht]{2.4 in}
        \includegraphics[width=\linewidth]{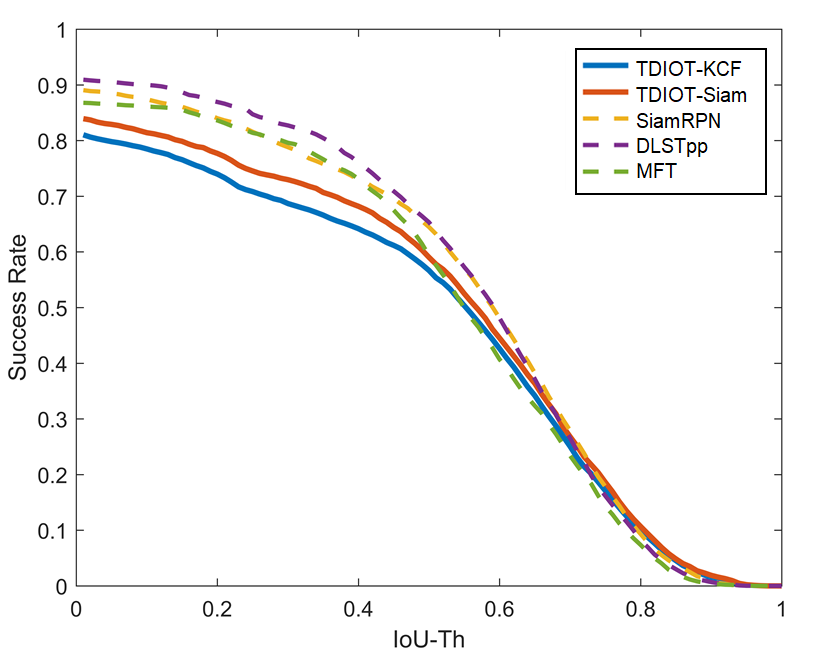}
        \caption{} 
    \end{subfigure}
     \caption{Success plots on VOT subsets (a) from VOT2016, (b) from VOT2018.}
    \label{SRPR}
\end{figure*}

We evaluated overall tracking performance of the proposed tracker against top trackers of VOT2018 and VOT2016 benchmarks, in particular,  SiamRPN \cite{siamRPN}, DLSTpp and MFT \cite{VOT2018} from VOT2018 and TCNN \cite{tcnn}, SSAT \cite{ssat} and MLDF \cite{MLDF} from VOT2016.
Four of these trackers; TCNN, SSAT, MLDF and SiamRPN employ deep neural networks, two of them; DLSTpp and MFT apply discriminative correlation filter.
As can be seen from the comparative success plot in Fig. \ref{SRPR} (a), TDIOT-Siam results in 3\% higher success rate compared to TCNN, the top tracker in VOT2016, at IoU-Th=0.5. And in VOT2018 videos, TDIOT-Siam has the comparable success rates especially at higher IoU-Th (Fig.\ref{SRPR}(b)) values. Performance of TDIOT-KCF is also reported for both datasets and evidently it can also achieve comparable results even though its overall performance is lower than TDIOT-Siam. We can conclude that, thanks to well localized proposals generated by SRPN, TDIOT is much more robust at higher IoU thresholds which indicates its high localization capability.
For more detailed analysis, we also report the attribute-based performance of TDIOT using accuracy and robustness, since the tracking performance varies depending on the attribute type. 
For this purpose, we used 5 attributes, which are employed for annotation of VOT videos, namely, illumination change, occlusion, motion change, camera motion and size change. 
Table \ref{table_sota_attr} and Table \ref{table_sota_attr8} demonstrate that the proposed tracker mostly provides the highest accuracy compared to the top trackers of VOT2016 and 2018, and comparable robustness in most of the cases. Specifically, TDIOT-Siam can outperform TCNN tracker by 4\% and 3\% on size and motion change attributes, respectively (Table \ref{table_sota_attr}). Furthermore, TDIOT-KCF provides the best robustness on illumination change and TDIOT-Siam has 4\% and 2\% better accuracy compared to SiamRPN on changes on illumination and size , respectively (Table \ref{table_sota_attr8}). We also report the average performance on both datasets, where TDIOT have better accuracy which indicates higher localization performance, and comparable but slightly worse robustness which indicates that the frequency of drifts off the target is high. These results confirm the effectiveness of inference architecture in handling such challenges through alleviating the problem caused by uniform anchors and applying a simple data association scheme.
\begin{table*}[ht!]
    \caption{ 
Attribute based results of TDIOT-KCF and TDIOT-Siam compared to the top trackers of VOT2016. The meanings of A, R, $\uparrow$ and $\downarrow$ are the same with the Table \ref{mrpnatt}. For each attribute the best score in terms of A and R is boldfaced.}
    \begin{center}
    \begin{tabular}{ c | l | c c c c c}
        \hline  
\small{Att}  &	&\footnotesize{TDIOT-KCF}&\footnotesize{TDIOT-Siam} &\footnotesize{TCNN \cite{tcnn}}&\footnotesize{SSAT \cite{ssat}} &\footnotesize{MLDF \cite{MLDF}}\\  \hline \hline
  \parbox[t]{1mm}{\multirow{2}{*}{\rotatebox[origin=c]{90}{\footnotesize{Illum}}}}   
  &  \small{A}$\uparrow$	&0.618 &{\bf 0.614}	&	0.599	&	0.594	&		0.575	\\ 
&  \small{R}$\downarrow$	&	{\bf 0.125}&	0.144&	0.206	&		0.184	&		{0.158}	

\\ 
 \hline 
   \parbox[t]{1mm}{\multirow{2}{*}{\rotatebox[origin=c]{90}{\footnotesize{Occ}}}} 
  &  \small{A}$\uparrow$	&0.587	&	0.589	&0.590	& {\bf 0.600}		&		0.580	\\ 
&  \small{R}$\downarrow$	&0.273	&0.263	&	{\bf 0.171}	&		0.245	&		0.224	

\\ 
 \hline
 \parbox[t]{1mm}{\multirow{2}{*}{\rotatebox[origin=c]{90}{\footnotesize{Mot}}}} 
   &  \small{A}$\uparrow$	&{\bf 0.610}	&{\bf 0.610}	&	0.582	&		0.579	&		0.521\\ 
&  \small{R}$\downarrow$	&0.251	&0.205&{\bf 0.179}	&		0.188	&		0.189	\\ 

 \hline
 \parbox[t]{1mm}{\multirow{2}{*}{\rotatebox[origin=c]{90}{\footnotesize{Cam}}}} 
    &  \small{A}$\uparrow$	&{\bf 0.621} &0.616	&		0.592	&		0.595	&		0.563		\\ 
&  \small{R}$\downarrow$	&0.186&	0.213&	0.117	&		0.138	&		{\bf 0.113}	

\\ 
 \hline
 \parbox[t]{1mm}{\multirow{2}{*}{\rotatebox[origin=c]{90}{\footnotesize{Size}}}} 
    &  \small{A}$\uparrow$	&{\bf 0.633}	&0.629	&	0.587	&		0.588	&		0.549		\\ 
&  \small{R}$\downarrow$	&0.202	&	0.190&	0.188	&	{\bf 0.166}	&		0.175	

\\ 
 \hline 
 \parbox[t]{1mm}{\multirow{2}{*}{\rotatebox[origin=c]{90}{\footnotesize{Avg.}}}} 
    &  \small{A}$\uparrow$	&{\bf 0.620}	&{\bf 0.620}	&	0.587	&		0.588	&		0.549		\\ 
&  \small{R}$\downarrow$	&0.220	&0.190&		0.188	&{\bf 0.166}	&		0.175	
\\ 
 \hline
  \end{tabular}
  \label{table_sota_attr}
  \end{center}
\end {table*}
 
\begin{figure*}[ht]
 \centering
  \begin{subfigure}[ht]{6.2 in}
   \includegraphics[width=\linewidth]{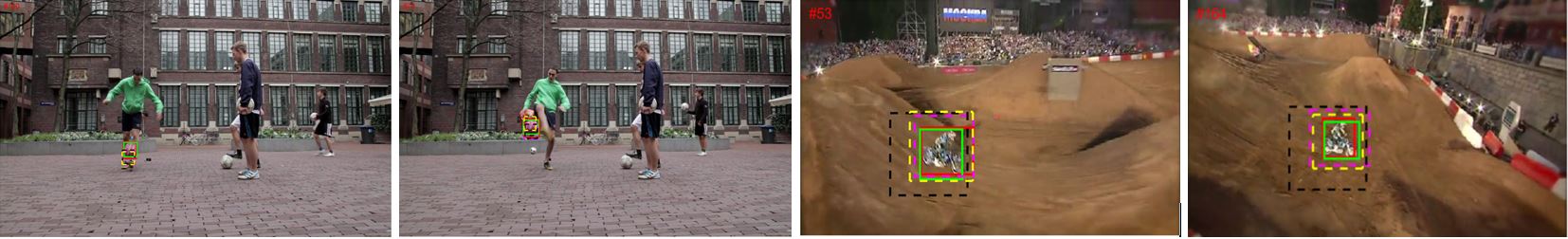}
    \caption{}
  \end{subfigure}
  \begin{subfigure}[ht]{6.2 in}
    \includegraphics[width=\linewidth]{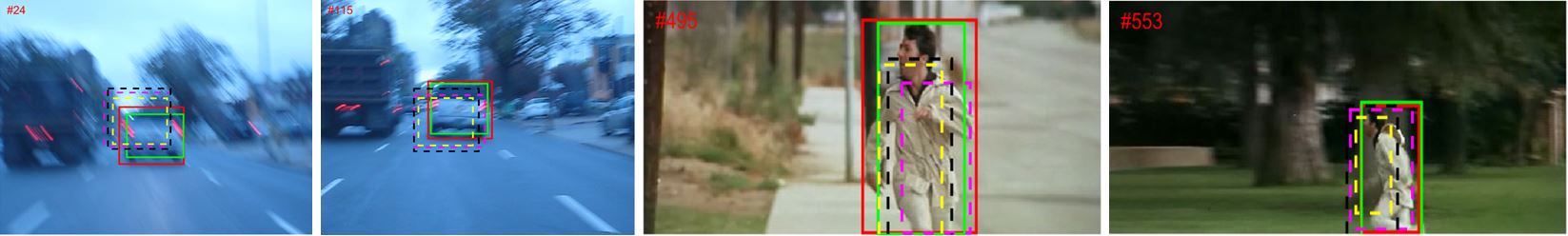}
   \caption{} 
  \end{subfigure}
    \caption{  Qualitative comparison of our method with state-of-the-art methods on VOT2016/2018 datasets. (a) ball and motocross1 video sequences. GT (green), TDIOT-Siam (red),  TCNN  (magenta), SSAT (yellow), MLDF(black), (b)car1 and graduate sequences. SiamRPN  (magenta), DLSTpp (yellow), MFT(black).  }
 \label{SOTAST}
\end{figure*}

  \begin{table*}[ht!]
\caption{ Attribute based of TDIOT-KCF and TDIOT-Siam compared to the top trackers of VOT2018. The meanings of A, R, $\uparrow$, $\downarrow$ are the same with the Table \ref{mrpnatt}. For each attribute the best score in terms of A and R is boldfaced.}
\begin{center}
  \begin{tabular}{ c | l | c c c c c   }
    \hline  

\small{Att}  &	&\footnotesize{TDIOT-KCF}&\footnotesize{TDIOT-Siam}&\footnotesize{SiamRPN \cite{siamRPN}}&\footnotesize{DLSTpp\cite{VOT2018}} &\footnotesize{MFT \cite{VOT2018}}\\  \hline \hline
  \parbox[t]{1mm}{\multirow{2}{*}{\rotatebox[origin=c]{90}{\footnotesize{Illum}}}}   
  &  \small{A}$\uparrow$&		{0.608}	& {\bf	0.620} &	0.581	&		0.580	&		0.567	\\ 
&  \small{R}$\downarrow$	&	{\bf 0.129}	&	0.155&	0.185	&	{0.143}	&		0.248 

\\ 
 \hline 
   \parbox[t]{1mm}{\multirow{2}{*}{\rotatebox[origin=c]{90}{\footnotesize{Occ}}}} 
  &  \small{A}$\uparrow$	&	0.573		&0.575&		0.598	&	{\bf 0.619}	&		0.598	\\ 
&  \small{R}$\downarrow$	&		0.306	&	0.296	&	0.175	&		{\bf 0.071}	&		0.133

\\ 
 \hline
 \parbox[t]{1mm}{\multirow{2}{*}{\rotatebox[origin=c]{90}{\footnotesize{Mot}}}} 
   &  \small{A}$\uparrow$	&	0.600	&0.601	&	0.593	&		{\bf 0.603}	&		0.578	\\ 
&  \small{R}$\downarrow$&		0.291	&	0.234&	0.159	&		{\bf 0.145}	&		0.192\\ 

 \hline
 \parbox[t]{1mm}{\multirow{2}{*}{\rotatebox[origin=c]{90}{\footnotesize{Cam}}}} 
    &  \small{A}$\uparrow$	&	{0.612}	&	0.607&	0.609	&	{\bf 0.615}	&		0.607	\\ 
&  \small{R}$\downarrow$	&		0.224	&0.237	&	0.170	&	{\bf 0.121}	&		0.163

\\ 
 \hline
 \parbox[t]{1mm}{\multirow{2}{*}{\rotatebox[origin=c]{90}{\footnotesize{Size}}}} 
    &  \small{A}$\uparrow$		&	 0.616	&{\bf 0.620}	&	0.601	&		0.588	&		0.569	\\ 
&  \small{R}$\downarrow$		&		0.249&0.226	&	{\bf 0.159}	&		0.167	&		0.225

\\ 
 \hline 
 \parbox[t]{1mm}{\multirow{2}{*}{\rotatebox[origin=c]{90}{\footnotesize{Avg.}}}} 
    &  \small{A}$\uparrow$	&		{\bf 0.610}	&{\bf 0.610}&		0.601	&		0.588	&		0.569	\\ 
&  \small{R}$\downarrow$	&		0.260	&0.220&		{\bf 0.159}	&		0.167	&		0.225
\\ 
 \hline
  \end{tabular}
  \label{table_sota_attr8}
  \end{center}
\end {table*}
In order to visualize differences from the existing trackers, we report the tracked object BBs on a number of video frames in Fig.\ref{SOTAST}. 
Four challenging video sequences, namely, \textit{ball1}, \textit{motocross1}, \textit{car1} and \textit{graduate} are chosen from VOT2016 and 2018 datasets. 
Fig.\ref{SOTAST}(a) illustrates the object BBs tracked by the proposed tracker TDIOT-Siam compared to TCNN, SSAT, and MLDF which are the top trackers of VOT2016. 
In \textit{ball1} sequence, all trackers catch the target object in both frames, but TDIOT-Siam provides higher localization accuracy. In \textit{motocross1} sequence, TDIOT can successfully handle size changes at frames 53 and 164, with the help of more qualified proposals generated by the proposed SRPN.
Fig.\ref{SOTAST}(b) illustrates improvement achieved by TDIOT-Siam on \textit{car1} and \textit{graduate} sequences compared to DLSTpp, SiamRPN and MFT where object tracking becomes harder because of blur and abrupt appearance changes. 
In \textit{car1} sequence, all state-of-the-art trackers have poor performance in handling blur, however TDIOT-Siam keeps tracking the target object by the help of local search and matching layer. 
As a results of TPS, changes in size and scale in \textit{graduate} sequence are successfully handled, with the help of the feedback from previous tracking results.

\subsection{ Comparison with State-of-the-Art Long-term Trackers}

We also evaluated TDIOT-KCF on videos from VOT-LT2018 dataset \cite{VOT2018} which is a specific benchmark for long term tracking. 
In this experiment, proposed method is compared against the top trackers of VOT-LT2018 benchmark in terms of  F-score, Precision (Pr) and Recall (Re) in Table\ref{SOTA2018}. 
The detailed comparisons shows that, proposed tracker achieves comparable performance to the state-of-the-art long-term trackers. 
In particular, TDIOT-KCF is at the $3^{th}$ place in terms of F-score which is the primary measure on the VOT-LT. Our tracker has 4\% higher Recall than LTSINT and 3\% lower than  DaSiam\_LT.

\begin{table}[ht!]
\caption{Performance comparison with trackers that participated in the VOT-LT2018 challenge along with their performance scores (F-score, Pr, Re) }
\begin{center}
  \begin{tabular}{ l  c  c c  }
    \hline  
{Tracker}  & F-score$\uparrow$	& Pr $\uparrow$ & Re$\uparrow$ \\  \hline \hline
 
  MBMD	&{\bf0.608} &{\bf0.636} &0.590 \\ 
  DaSiam\_LT	& 0.599 & 0.606 & {\bf0.601}\\
  TDIOT KCF	&0.572&0.584 &0.569	\\ 
  TDIOT Siam	&0.542&0.562 &0.529	\\ 
  MMLT	& 0.545& 0.570&0.530 \\
  LTSINT&0.540 &0.536 &0.548 \\
  SYT&0.517 &0.526 &0.511 \\
  SLT&0.448 &0.463 &0.439 \\
  ASMS&0.290 &0.370 &0.260 \\
  FuCoLot&0.17 & 0.597& 0.105\\
 \hline 
  \end{tabular}
  \label{SOTA2018}
  \end{center}
\end {table}

\begin{figure*}[ht]
\includegraphics[width=18cm]{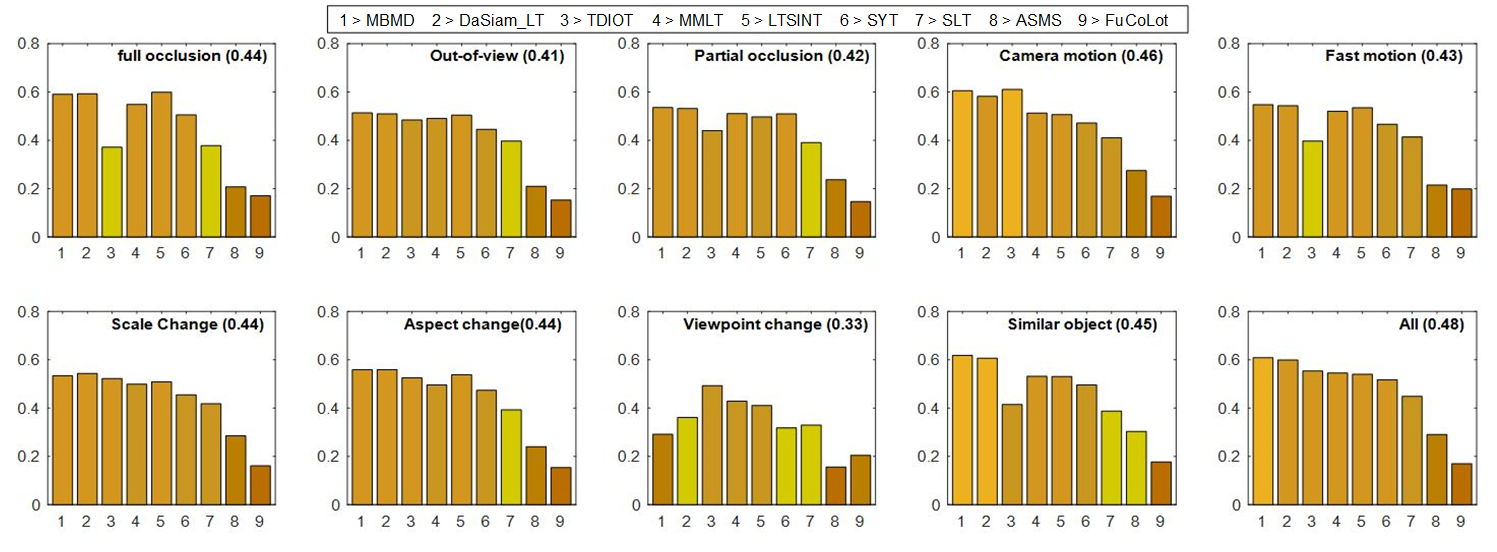}
 \centering
 \caption{Maximum F-score for the visual attributes compared to VOT-LT2018 state-of-the-arts.}
 \label{attbased2018}
\end{figure*}

For detailed performance analysis on long term sequences, we also report the results on 9 challenging attributes.
Fig.\ref{attbased2018} demonstrates that our tracker can effectively handle some of the challenges including camera motion and viewpoint change. 
Reported results indicate that TDIOT-KCF outperforms all other trackers in viewpoint change attribute.

 \begin{figure*}[ht]
 \centering
  \begin{subfigure}[ht]{6.15 in}
   \includegraphics[width=\linewidth]{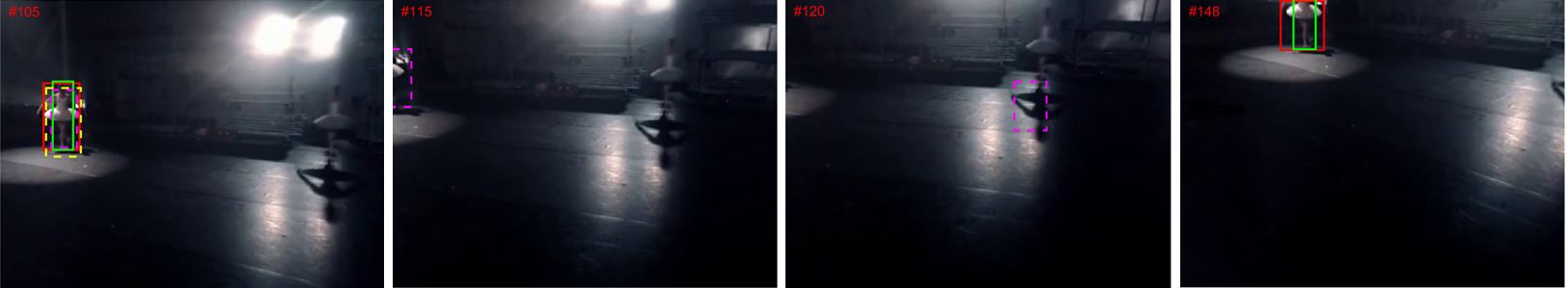}
    \caption{}
  \end{subfigure}
  \begin{subfigure}[ht]{6.15 in}
    \includegraphics[width=\linewidth]{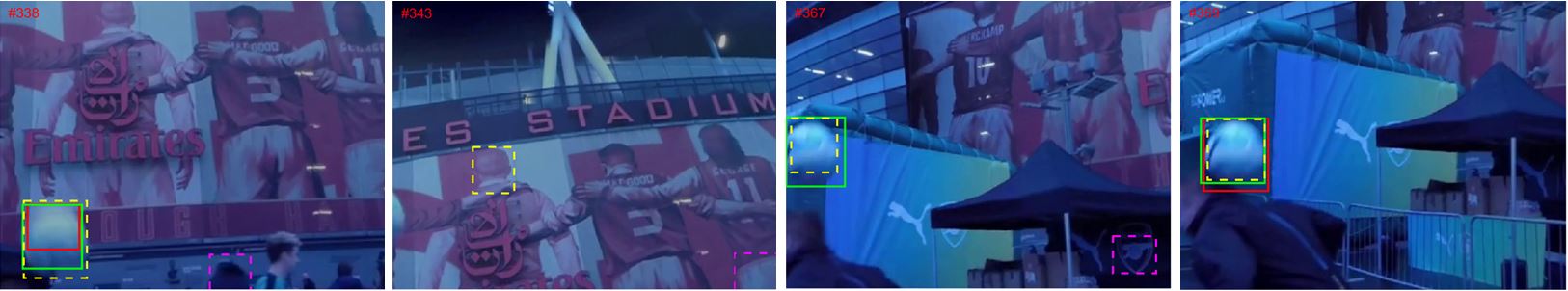}
   \caption{} 
  \end{subfigure}
    \caption{Qualitative comparison of our method with state-of-the-art methods. (a) ballet video sequence (b) freestyle video sequences. GT (green), TDIOT-KCF (red), MBMD (yellow), DaSiam\_LT (magenta).}
 \label{SOTALT}
\end{figure*}

To visualize the performance of our LT tracker, we provide comparative results of our tracker and top 2 methods reported in VOT-LT2018 benchmark \cite{VOT2018}. 
If the object is out-of-view, ground truth (GT, green BB) is not drawn in the frame. 
Also, if the tracker concludes the target has disappeared from the frame according to prediction certainty score, related BB is not drawn in the frame either. 
As shown in Fig. \ref{SOTALT}(a), all trackers manage to catch the target object in frame 105.  
However, DaSiam\_LT fails after object disappears in frame 115, where TDIOT detects disappearance of the target with a simple thresholding after a 1 frame delay. 
In the last frame, with the help of local-to-global search scheme, TDIOT-KCF is the only tracker that manages to continue tracking after the target object re-appears at a different location. 
In \textit{freestyle} sequence (Fig. \ref{SOTALT}(b)), although MBMD provides higher localization accuracy in frame 338, our method can detect the target disappearance successfully when the MBMD drifts away to another similar object easily. 
After object re-entrance, both MBMD and TDIOT-KCF detects the object with 2 and 4 frame delays, respectively. 
It is important to note that, MBMD tracker uses MDNet, a deep network which has stronger discriminative power, to verify the object exit/entrance, while we use a simpler scheme that  discriminates the LBP histogram with chi-square distance metric.

In addition to conventional metrics mentioned above, we also  examined the performance of the TV layer. 
Firstly, we report the percentage of times the tracker successfully detects object disappearance or re-appearance. 
In particular, TDIOT-KCF detects 71\% of the disappearance, furthermore 21\% of them occur without any delay. 
On the other hand, the percentage of re-appearance detection is 56\%, less than the other case, but this time 36\% of them occur as soon as object appears in the frame. 
Since the tracker can detect disappearance and re-appearance before or after it occurs, we also reported the average number of frames required to detect disappearance and re-appearance.
Specifically, TDIOT-KCF can detect the disappearances within frame range [-5:10], while reappearances are detected within [-21:5]. Here, negative sign indicates that the tracker confirms the disappearances/appearance before it occurs.

The   source   code   and   TDIOT output videos  are   accessible   at \url{https://github.com/msprITU/TDIOT}. 

\section{Conclusions}

We have presented an inference framework, TDIOT, to convert a visual object detector to a video object tracker by replacing the inference level proposal generation and head layers of the baseline detector. 
Our contributions are at the inference phase thus the proposed framework does not require any specific training for tracking purpose and the pre-trained visual object detector is sufficient to attain video object tracking goal. It is shown that the introduced proposal generation network referred as SRPN has various advantages compared to conventional RPN models. It grants target-awareness to the proposal generation scheme and allows arbitrary shaped anchor generation which enables TDIOT to keep tracking especially when high motion or scale changes are present. 

In order to alleviate miss detections where the tracker fails to identify any objects in the frame, we employed a SiamFC based local search and matching scheme to decrease temporal discontinuities.
 TDIOT is adopted to long-term tracking by adding an effective local-to-global search strategy based on LBP histogram similarity for verification of the target object. Our evaluations demonstrate that accuracy has slightly increased when SiamFC is replaced by KCF in long term tracking. Extensive experiments on short term and long term tracking video sequences denote that TDIOT achieves favorable performance against state-of-the-art trackers.

In order to achieve high detection rates, we used Mask R-CNN trained on COCO dataset as a baseline detector. It is straightforward to replace it by a one-pass deep object detector to speed up tracking, with the expense of lower accuracy. Moreover,  it is possible to improve the tracking performance by pre-training the detector on a high resolution dataset \cite{peddet}.

\ifCLASSOPTIONcaptionsoff
  \newpage
\fi

\appendices

\bibliographystyle{IEEEtran}

\bibliography{myBib}

\vspace{-4.5 cm}
\vspace{-5 cm}
\vspace{-5 cm}

\end{document}